\documentclass[sigconf]{acmart}

\AtBeginDocument{%
  \providecommand\BibTeX{{%
    \normalfont B\kern-0.5em{\scshape i\kern-0.25em b}\kern-0.8em\TeX}}}

\usepackage{booktabs} 

\usepackage{wrapfig}
\usepackage{sidecap}
\usepackage{graphicx}
\usepackage{subcaption}
\usepackage{amsfonts}

\usepackage{amsmath,xparse,bm}
\usepackage{graphicx}
\usepackage[ruled]{algorithm2e}

\newcommand\method{PoMS }

\newcommand{\argmin}{\mathop{\mathrm{argmin}}\limits}

\copyrightyear{2021}
\acmYear{2021}
\setcopyright{acmlicensed}
\acmConference[GECCO '21]{2021 Genetic and Evolutionary Computation Conference}{July 10--14, 2021}{Lille, France}
\acmPrice{15.00}
\acmDOI{10.1145/3449639.3459320}
\acmISBN{978-1-4503-8350-9/21/07}
\acmBooktitle{2021 Genetic and Evolutionary Computation Conference (GECCO '21), July 10--14, 2021, Lille, France}

\begin{document}

\title{Policy Manifold Search}
\subtitle{Exploring the Manifold Hypothesis for Diversity-based Neuroevolution}


\author{Nemanja Rakicevic}
\orcid{1234-5678-9012}
\affiliation{%
  \institution{Imperial College London}
}
\email{n.rakicevic@imperial.ac.uk}

\author{Antoine Cully}
\orcid{1234-5678-9012}
\affiliation{%
  \institution{Imperial College London}
}
\email{a.cully@imperial.ac.uk}

\author{Petar Kormushev}
\orcid{1234-5678-9012}
\affiliation{%
  \institution{Imperial College London}
}
\email{p.kormushev@imperial.ac.uk}

\renewcommand{\shortauthors}{N. Rakicevic, A. Cully and P. Kormushev}

\begin{abstract}
    %
    Neuroevolution is an alternative to gradient-based optimisation that has the potential to avoid local minima and allows parallelisation. The main limiting factor is that usually it does not scale well with parameter space dimensionality. Inspired by recent work examining neural network intrinsic dimension and loss landscapes, we hypothesise that there exists a low-dimensional manifold, embedded in the policy network parameter space, around which a high-density of diverse and useful policies are located. This paper proposes a novel method for diversity-based policy search via Neuroevolution, that leverages learned representations of the policy network parameters, by performing policy search in this learned representation space. Our method relies on the Quality-Diversity (QD) framework which provides a principled approach to policy search, and maintains a collection of diverse policies, used as a dataset for learning policy representations. Further, we use the Jacobian of the inverse-mapping function to guide the search in the representation space. This ensures that the generated samples remain in the high-density regions, after mapping back to the original space. Finally, we evaluate our contributions on four continuous-control tasks in simulated environments, and compare to diversity-based baselines. 
\end{abstract}

\ccsdesc[500]{Computing methodologies~Dimensionality reduction and manifold learning}
\ccsdesc[500]{Computing methodologies~Evolutionary robotics}
\ccsdesc[300]{Computing methodologies~Neural networks}

\keywords{Neuroevolution, Manifold Learning, Quality-Diversity, MAP-Elites} %

\maketitle


\vspace{-0.15cm}

\section{Introduction}

    \begin{figure}[!t]
    \centering
        \includegraphics[width=1\linewidth]{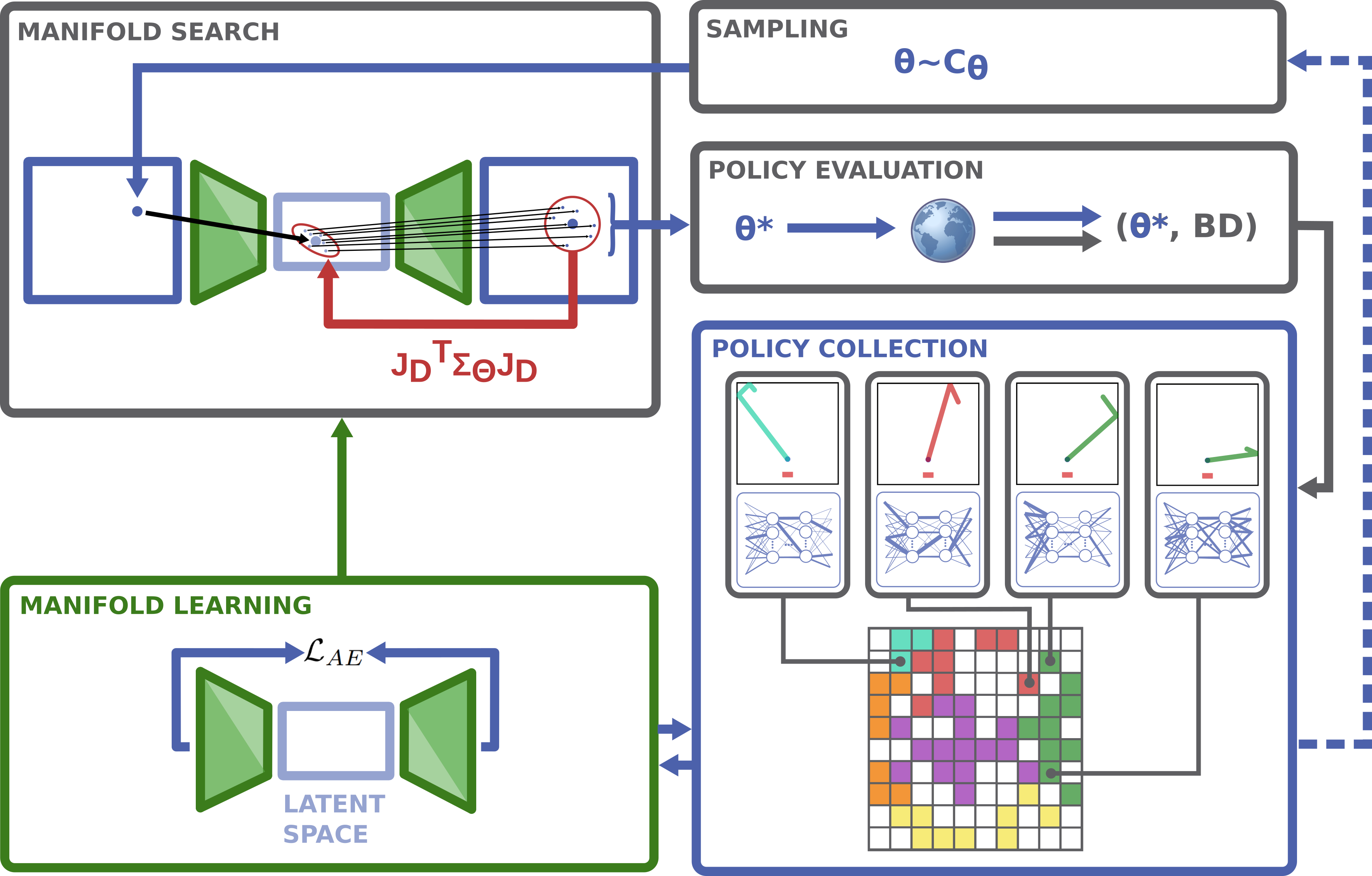}
        \caption{
            The proposed Policy Manifold Search approach components.
            Grey boxes at the top are used within the \textit{manifold search} phase which adds policies to the collection,
            while the \textit{manifold learning} is shown in green and uses all the data from the policy collection ($\mathrm{C}_{\bm{\theta}}$).
            Blue arrows show policy network parameter ($\theta$) flow,
            grey arrows evaluation information ($\theta^{*}$, BD),
            green arrow is the current autoencoder parameters, and
            red arrow depicts decoder Jacobian scaling.
        } 
        \label{fig:schematics}
    \end{figure}
    
    In recent years, we have seen significant progress in tackling continuous control tasks by gradient-free approaches based on Evolutionary Algorithms, such as Neuroevolution \cite{salimans2017evolution, such2017deep, stanley2019designing}.
    Compared to Deep Reinforcement Learning (RL) approaches relying on gradient-based policy network optimisation \cite{schulman2017proximal}, Neuroevolution achieves similar performance while allowing for greater parallelisation.
    %
    %
    %
    %
    %
    However, these performance-based methods aim to find a unique set of controller parameters, although there are cases where a diversity of solutions is needed and this becomes an issue.
    An example includes environments that can accommodate multiple tasks, and that different controller policies are necessary to solve and adapt to a dramatically changing environment or recover from damage~\cite{cully2015robots}.
    %
    %
    Quality-Diversity (QD) framework~\cite{pugh2016quality, cully2017quality}, has been introduced as an elegant approach to maintaining a diversity of controller policies.
    A popular QD approach, MAP-Elites~\cite{mouret2015illuminating}, maintains a diversity of policies by having each distinct policy assigned to the policy collection based on the \textit{behaviour descriptor}, i.e. low-dimensional representation of the corresponding policy's behaviour in the environment~\cite{cully2017quality}. 
    New solutions are then generated by modifying existing ones via \textit{mutation operators}.
    Applications of MAP-Elites are often limited to low-dimensional or open-loop controllers~\cite{mouret2015illuminating}, while Deep RL and Neuroevolution have shown to be capable of learning deep neural networks for continuous control, increasing the agent's capabilities.
    Still, the curse of dimensionality makes evolution quite challenging, particularly in MAP-Elites, and prevents it to scale with the parameter dimensionality~\cite{colas2020scaling}.
    A popular approach to addressing this issue is representation learning~\cite{cayton2005algorithms, bengio2013representation}, which has been studied extensively in the context of input spaces such as environment observation vectors, images, graphs etc~\cite{hausman2018transferable, pere2018unsupervised}. 
    Representation learning relies on the \textit{manifold hypothesis} which states that high-dimensional data tends to lie in the vicinity of a low dimensional manifold.
    However, this notion has not been investigated much in the context of neural network parameter spaces.
    Recent work on examining neural network parameters space properties, gives us insights on the possible existence of low-dimensional manifolds in the network parameter space~\cite{draxler2018essentially, garipov2018loss, li2018measuring, richemond2019biologically, frankle2019lottery, morcos2019one}.
    %
    %
    %
    Moreover, using a notion of a manifold via the hypervolume of elites within MAP-Elites, showed promising results~\cite{vassiliades2018discovering}.
    These findings suggest a natural question whether there exists a manifold embedded in the high-dimensional network parameter space, and what are the potential benefits for policy search via Neuroevolution using MAP-Elites?
    Inspired by the \textit{manifold hypothesis},
    we hypothesise that there exists a lower-dimensional, non-linear manifold, embedded in the high-dimensional policy network parameter space, which contains a high density of solutions for a particular task.
    In this paper, we propose a novel approach, Policy Manifold Search (PoMS), that learns a mapping to a manifold in the policy network parameter space, which is used for Neuroevolution-based policy search.
    We expand the MAP-Elites framework, by using its policy collection as a dataset for manifold learning via an Autoencoder (AE).
    This is an important component as the quality of the dataset dictates the quality and generalisability of the learned latent representations.
    Focusing the search in the manifold with a high solution density, improves the sample complexity and the diversity of discovered behaviours.
    When mapping back from the manifold to the original parameter space, we take into account the additional transformation incurred, due to the imperfect mapping of the AE's decoder function.
    This leads to a consistent parameter search regardless of the local structure of the latent representation, by accounting for the distortions induced by the decoder.
    Finally, we focus solely on diversity search and do not use performance as a selection criteria.

    %
    In order to evaluate our hypothesis, we conduct ablation studies of the algorithm components, as well as comparisons to the state-of-the-art MAP-Elites approaches.
    These experiments are executed on four different continuous control tasks.
    As a performance indicator, we use the \textit{behaviour coverage} metric, typically used in QD literature~\cite{cully2017quality}, to quantify the diversity of policies achieved within the policy collection.
    The results suggest that learning a policy manifold explicitly and using it for policy search works well, but an additional regularisation of the search process using the decoder Jacobian is necessary in order to have a well-behaved policy search.



\vspace{-0.05cm}

\section{Related Work}
\label{sec:related_work}
    
    In this section, we present relevant work on policy behaviour diversity, as well as parameter representation and generation. 
    
    \textbf{Manifold Learning} 
        aims to obtain a lower-dimensional representation, i.e. a manifold, embedded in the high-dimensional input space. 
        It is based on the \textit{manifold hypothesis} that assumes a high density of useful datapoints located in the vicinity of such a manifold.
        This notion has been thoroughly explored under \textit{representation learning}~\cite{bengio2013representation},
        and applied in RL~\cite{chen2018active, eysenbach2018diversity, rusu2018meta},
        with important insights on how to exploit the structure of the manifold to improve an algorithm's robustness~\cite{rifai2011contractive, rifai2012generative}.
        However, this body of work has focused on manifold learning in the context of input spaces such as images, environment observations, etc, which have different structural properties compared to neural network parameter spaces.
        Recent work examines the concept of a manifold in the context of network parameter spaces, and show that the local optima of deep networks can be connected by continuous paths which lie in low loss value regions of the network parameter space~\cite{draxler2018essentially, garipov2018loss, freeman2016topology}.
        This is related to the redundant nature of networks as multiple configurations could lead to the same behaviour~\cite{li2015convergent}.
        Moreover, there are some insights on the existence of an \textit{intrinsic dimension} of the network parameter space, that is dependant on the particular task's complexity and is sufficient for solving such a task~\cite{denil2013predicting, yang2015deep, li2018measuring, richemond2019biologically, gaier2019weight}.
        %
        Another line of work investigates the \textit{lottery ticket hypothesis}, which states that deep networks contain subnetworks which can reach the same performance as the original network, when trained in isolation~\cite{frankle2019lottery, morcos2019one}.
        These findings give us inspiration that such a manifold could be found and used for efficient policy search in QD.

    \textbf{Parameter-generating networks} 
        have been proposed as an approach to few-shot adaptation of the (controller) network parameters, conditioned on the observations or latent variables as inputs, in the context of fast/slow weights~\cite{gomez2005evolving, ba2016using}, meta-learning~\cite{rusu2018meta}, dynamic filter networks~\cite{jia2016dynamic} or hypernetworks~\cite{ha2016hypernetworks}.
        %
        %
        The above-mentioned approaches do not explicitly learn a parameter manifold from which the controller network parameters are generated, rather a differentiable mapping from the observation space which can be updated via gradient descent.
        Conversely, in recent work a manifold is learned from a collection of converged neural network policies~\cite{chang2019agent}, or simple motion primitive based controllers~\cite{jegorova2018generative}.
        They use learned representations for analysing and assessing new generated policy network parameters, rather than for policy search.
    	To the best of our knowledge, Chang et al \cite{chang2019agent} is the only work besides ours, that learns policy network parameter representation.
    	However, they use a set of fully converged policy networks to train the AE, which is costly to produce and could lead to a less generalisable representation.

    \textbf{Quality Diversity}
        (QD) algorithms \cite{pugh2016quality, cully2017quality} have been recently introduced as a framework that generates a collection of high-performing and diverse solutions. 
        Most popular approaches are Multi-dimensional Archive of Phenotypic Elites (MAP-Elites) \cite{mouret2015illuminating, cully2015robots} and Novelty Search with Local Competition \cite{lehman2011evolving},
        which differ in how they select and maintain a collection of controller policies based on their exhibited behaviours.
        In this work, we expand on MAP-Elites, due to its simplicity of implementation and proven performance in various applications \cite{gravina2019procedural, cully2015robots, urquhart2019illumination}.
        Although MAP-Elites usually uses simple low-dimensional parameterised controllers~\cite{cully2015robots},
        recent work proposed scaling MAP-Elites to Neuroevolution and applied it to more complex environments \cite{colas2020scaling}, by
        combining MAP-Elites with Evolutionary Strategies to effectively search the high-dimensional network parameter space.
        Two recently introduced MAP-Elites-based approaches present ideas which are close to the notion of controller parameter manifolds.
        MAP-Elites via Elite Hypervolumes~\cite{vassiliades2018discovering} uses a mix of isotropic and directional Gaussian operators allowing for an adaptive search that implicitly explores the hypervolume of the elites, which is similar to the learned manifold in PoMS, but in the original parameter space.
        Data-driven encoding MAP-Elites~\cite{gaier2020discovering}, combines the line mutation with a reconstruction-crossover operator, based on an AE trained on the policies contained in the collection, similarly to our approach.
        %
    	%
    	%
    	%
    %
    %
    
    \textbf{Task-conditioned policies}
        are a widely used framework for achieving a diversity of policies in RL.
        These policies are conditioned on a task context identifier, sampled from a discrete \cite{hausman2018transferable, eysenbach2018diversity} or continuous \cite{sharma2019dynamics} distribution.
        %
        %
        Diversity is All You Need (DIAYN) \cite{eysenbach2018diversity} focuses purely on discovering diverse skills, i.e. behaviours, and does not consider extrinsic environment rewards.
        %
        %
        %
        While QD and task-conditioned policies both aim to promote behaviour diversity, they differ in how the diversity is maintained and in the definition of skills, i.e. behaviours.
        QD requires certain level of domain knowledge, still, it is able to consider a significantly larger number of diverse behaviours.
        %
        %
        %
        %


\section{Policy Manifold Search}
\label{sec:method}
    
    In this section, we describe the proposed Policy Manifold Search (PoMS) algorithm (Fig \ref{fig:schematics}).
    %
    \method is an iterative algorithm, where each iteration consists of the \textit{parameter manifold learning} and \textit{parameter manifold search} phases (Algorithm \ref{algo:poms}).
    The former consists of finding a latent representation of the parameter space and the corresponding mapping functions, while the latter performs search in the learned latent space to generate new diverse policies.
    Each iteration potentially adds new policies to the collection, following the MAP-Elites framework~\cite{cully2015robots}, which are then used to refine the latent representation in the next iteration.
    The goal of PoMS is to focus the search in the manifold which contains a high density of solutions, in order to improve the sample complexity and the behaviour coverage.

    \subsection{Preliminaries}
        
        Let us consider a typical RL setting, with an environment defined as a Markov Decision Process (MDP) $(\mathcal{S}, \mathcal{A}, \mathcal{T}, \mathcal{R}, \gamma)$,
        with the state space $\mathcal{S}$, action space $\mathcal{A}$, and the deterministic transition function $\mathcal{T}:\mathcal{S} \times \mathcal{A} \rightarrow \mathcal{S}$.
        The reward function $\mathcal{R}$ and discount factor and $\gamma$ are not used in this work.
        We define a deterministic policy $\pi_{\theta}$, parameterised as a deep neural network, that maps the current state to the action to be taken at that state $a_t = \pi_{\theta}(s_t)$.
        The policy parameters $\theta$ are a P-dimensional set of network weights and biases, $\theta \in \mathbb{R}^{P}$, such that each point in the policy parameter space defines a unique policy.
        %
        %
        During an episode of length $T$, an agent interacts with the environment, using the policy $\pi_{\theta}$, thus generating a trajectory $\tau = \{s_i, a_i\}_1^T$.
        We want to distinguish how a certain deterministic policy $\pi_{\theta_k}$ interacts with the environment in a quantifiable way.
        To this end, we use the concept of a \textit{Behaviour Descriptor} (BD) from the QD literature~\cite{cully2017quality} which aims to uniquely describe an episode rollout.
        The BD is formalized as a mapping from a state-trajectory $\tau$ space, to a $b$-dimensional behaviour space $BD: \mathrm{T} \rightarrow \mathcal{B}$.
        
        The MAP-Elites framework maintains a collection of multiple policy parameters, in a multi-dimensional cell-grid $\mathrm{C}_{\bm{\theta}}$, which is indexed by the behaviour index $b \in \mathcal{B}$ obtained using the BD.
        Different policies can produce the same $b$, but a specific deterministic policy will map to a unique behaviour (surjective mapping).
        To resolve the surjective mapping, usually each cell of the grid is populated by a single highest-performing policy based on some performance metric.
        %
        The aim of MAP-Elites is to fill all the cells of $\mathrm{C}_{\bm{\theta}}$ with the best possible policies through an iterative process. 
        Each iteration consists of (i) randomly selecting a batch of individuals from the collection, (ii) applying a mutation and evaluating these modified individuals, (iii) based on the outcome of the evaluation, add the new individuals to the grid if the corresponding cell is vacant or if they outperform the currently occupying individual.
        In this study, we focus solely on policy behaviour diversity, so a cell occupancy conflict is resolved by a coin-flip.
        A typical mutation operator adds an isotropic Gaussian noise $\mathcal{N}(0, \Sigma_{\Theta})$, with a unit covariance matrix $\Sigma_{\Theta} = \sigma_{\Theta}\mathbb{I}$, where $\sigma_{\Theta}$ is a hyperparameter.
    
      \vspace{-0.2cm}

        \begin{algorithm}
    \small
        \caption{Policy Manifold Search (PoMS)}
        \label{algo:poms}
        \SetAlgoLined
        \DontPrintSemicolon
        \tcp{initialisation}
        $\xi \leftarrow \text{Xavier-Glorot}$ \;
        $\mathrm{C}_{\bm{\theta}} = \emptyset$ \;
        $\bm{\theta}^{INIT} \sim \mathcal{U}(-1,1)$ \;
        $\bm{\tau}^{INIT} = \text{environment\_eval}(\pi_{\bm{\theta}^{INIT}})$ \;
        $\bm{b}^{INIT} = \mathcal{BD}(\bm{\tau}^{INIT})$ \;
        $\mathrm{C}_{\bm{\theta}}[\bm{b}^{INIT}] \leftarrow \bm{\theta}^{INIT}$ \;
        \For{nloop \textbf{in} PoMS\_loops}{
            \tcp{parameter manifold search phase}
            \For{niter \textbf{in} MAP-Elites\_iterations}{
                $\bm{\theta}^{SEL} \sim \mathrm{C}_{\bm{\theta}}$ \;
                $\bm{\theta}^{MUT} = \text{region\_based\_search}(\bm{\theta}^{SEL})$ \;
                $\bm{\tau}^{MUT} = \text{environment\_eval}(\pi_{\bm{\theta}^{MUT}})$ \;
                $\bm{b}^{MUT} = \mathcal{BD}(\bm{\tau}^{MUT})$ \;
                $\mathrm{C}_{\bm{\theta}}[\bm{b}^{MUT}] \leftarrow \bm{\theta}^{MUT}$ \;
            }
            \tcp{parameter manifold learning phase}
            \For{ $\bm{\theta}_{\text{batch}}$ \textbf{in} $\mathrm{C}_{\bm{\theta}}$}{
                $\hat{\bm{\theta}}_{\text{batch}} = f_{D} \circ f_{E}(\bm{\theta}_{\text{batch}}; \xi)$ \;
                $\mathcal{L}_{AE} \equiv \| \bm{\theta}_{\text{batch}} -  \hat{\bm{\theta}}_{\text{batch}} \|_2^2$ \;
                $\argmin_{\xi} \mathcal{L}_{AE}$ \;
            }
        }
    \end{algorithm}
    %
    %
    %
    \vspace{-0.56cm}

    \subsection{Parameter Manifold Learning Phase}
        The main insight of \method is learning a lower-dimensional manifold $\mathbb{R}^{M}$, embedded in $\mathbb{R}^{P}$ where $M<<P$, around which a high-density of interesting (i.e. non-degenerate) policies are located.
        This manifold can then serve as a smaller search space for a more efficient exploration.
        To find the manifold, we first generate a set of randomly distributed policy network parameters (uniformly sampled $\theta_i \sim \mathcal{U}(-1,1)$) and add them to the policy collection according to the MAP-Elites framework, which preserves diverse and unique behaviours.
        Then, we use the parameters that have been added to the policy collection to train a dimensionality reduction algorithm, like a deep AutoEncoder (AE) \cite{kramer1991nonlinear, hinton2006reducing}.
        The bottleneck layer of the AE defines the latent parameter representation space, which we refer to as the learned manifold.
        In this way, each point in the original parameter space $\theta_i$ can be directly mapped into the corresponding point on the manifold $z_i  \in \mathbb{R}^{M}$ using the encoder ($f_{E}$), and reconstructed back using the decoder ($f_{D}$).
        The AE used is fully-connected and symmetrical, where $f_{E}$ and $f_{D}$ are parameterised by $\xi = \{ \xi_E, \xi_D\}$.
        The AE parameters $\xi$ are learned by minimising the reconstruction loss:
        $\mathcal{L}_{AE} = \frac{1}{|\mathrm{C}_{\bm{\theta}}|}\sum_i^{|\mathrm{C}_{\bm{\theta}}|}  \| \theta_i -  \hat{\theta}_i  \|_2^2 $,
        %
        %
        where $\hat{\theta}_i = f_{D} \circ f_{E}(\theta_i; \xi)$.
        In contrast with recent works \cite{chang2019agent}, we do not apply any additional regularisation on the latent space, and train the AE in an unsupervised manner to reconstruct its inputs. 
        As opposed to the common training strategy on static datasets, in the case of \method, it is not beneficial to normalise the training data before fitting the AE, as periodic additions to the collection lead to instability in the training.
        
        Each iteration of \method is likely to add new policies to the collection.
        The \textit{parameter manifold learning} phase then uses all the policy parameters from the collection and continues the training of the AE to refine the latent representation.
        Further details on the training procedure and hyperparameters, are presented in Appendix~\ref{appendix:training}.

    \subsection{Parameter Manifold Search Phase}
        %
        %
        One of the strengths of MAP-Elites is that it constantly applies small mutations to the ``elites", a subset of solutions contained in the collection, therefore only exploring around solutions that have shown to be either high-performing (i.e. exploiting) or different (i.e.  exploring).
        In PoMS, we sample a random subset of solutions from the policy collection, as we focus on behaviour diversity rather than performance.
        Further, we also apply small mutations to the selected solutions, although these mutations are applied to the latent representations of the solutions in the latent space, as opposed to directly in parameter space.
        %
        %
        However, a small perturbation in the latent space can lead to a very large perturbation in the parameter space, once reconstructed, due to the complexity of the learned decoder.
        Therefore, there is a significant risk that applying mutations directly in the latent space (e.g. via Gaussian noise) and reconstructing, will lead to an uncontrolled mutation overall, similar to random search.  
        %
        %
        %
        %
        %
        %
        %
        %
        %
        %
        
        \textbf{Considering the decoder Jacobian.}
        %
        %
        To address this issue, we propose limiting the magnitude of the reconstructed samples, by making the latent parameter space search heteroscedastic.
        %
        We impose that each mutated point in the latent space ${z}' \sim \mathcal{N}(z_k, \Sigma_{Z})$, when reconstructed, lands within an isotropic Gaussian distribution in the original parameter space, i.e. $f_{D}(z') \sim \mathcal{N}(f_{D}(z_k), \Sigma_{\Theta}=\sigma_{\Theta}\mathbb{I})$.
        Meaning, the reconstructed samples are limited to a hyper-sphere region centered around the selected policy parameters.
        To achieve this, we use the Jacobian of the decoder, which gives us a linear approximation (first order Taylor expansion) of the transformation around a specific point in the latent space $z$, denoted as $\mathbf{J}_{D}(z)$: 
        \begin{equation}
            \mathbf{J}_{D}(z) = J_{D}(z)_{ij} = \frac{\partial f_{D}(z)_i}{\partial z_j}
        \end{equation}
        %
        %
        %
        %
        %
        %
        %
        %
        %
        %
        %
        %
        Let us define the desired covariance matrix of an isotropic Gaussian in the original parameter space as $\Sigma_{\Theta} = \sigma_{\Theta}\mathbb{I}$, where $\sigma_{\Theta}$ is the desired radius of the spherical Gaussian and $\mathbb{I} \in \mathbb{R}^{P\times P}$ is the square identity matrix. The objective is to estimate the appropriate covariance matrix for the Gaussian noise $\Sigma_{Z}$ applied in the latent space as a function of $\Sigma_{\Theta}$ and the Jacobian. This is given by:
        \begin{equation}
        \label{eq:sigma}
            \Sigma_{Z} = \mathbf{J}_{D}^T \Sigma_{\Theta} \mathbf{J}_{D}
        \end{equation}
        %
        %
        For the full derivation of Equation (\ref{eq:sigma}) using Taylor expansion, please refer to Appendix~\ref{appendix:jacobian}.
            \begin{algorithm} 
            \small
            \caption{Region-based search}
            \label{algo:mix_region}
            \SetAlgoLined
            \DontPrintSemicolon
            \textbf{Input:} $\bm{\theta}^{SEL}$; $\mathrm{C}_{\bm{\theta}}$, $\Sigma_{\Theta}$, $f_{E}$, $f_{D}$ \;
            \textbf{Output:} $\bm{\theta}^{MUT}$  \;
                $\epsilon_{recn} = \frac{1}{|\mathrm{C}_{\bm{\theta}}|}\sum_i^{|\mathrm{C}_{\bm{\theta}}|}  \| \theta_i -  \hat{\theta}_i  \|_2^2 $ \;
                    $\bm{\theta}^{MUT} = \emptyset$\;
                    \For{$\theta_i$ \textbf{in} $\bm{\theta}^{SEL}$}{
                        $\hat{\theta}_i = f_{D} \circ f_{E}(\theta_i)$ \;
                        \eIf{$ \| \theta_i - \hat{\theta}_i \|_2^2 < \epsilon_{recn} $}{
                            $z_i = f_{E}(\theta_i)$ \;        
                            $\mathbf{J}_{D} = \text{calculate\_jacobian\_matrix}(f_{D}; z_i)$ \;
                            $\Sigma_{Z} = \mathbf{J}_{D}^T \Sigma_{\Theta} \mathbf{J}_{D}$ \;
                            $z_i^{MUT} = z_i + \mathcal{N}(0, \Sigma_{Z})$ \;
                            $\theta_i^{MUT} = f_{D}(z_i^{MUT})$ \;
                        }{
                            $\theta_i^{MUT} = \theta_i + \mathcal{N}(0, \Sigma_{\Theta})$ \;
                        }
                        $\bm{\theta}^{MUT}  \leftarrow \theta_i^{MUT}$\;
                    }
            \end{algorithm}
        %
        %
        
        \textbf{Region-based mixing strategies.}
        %
        %
        Performing the parameter search in the latent space has the advantage of offering a smaller search space with a high-density of different and interesting policies.
        However, like in most autoregressive algorithms, the AE is unable to generalise far beyond the training set data support.
        To overcome this problem, we employ a region-based exploration strategy. 
        %
        If the reconstruction error of a selected $\theta_i$ is below a threshold value $\epsilon_{recn}$, we perform the mutation in the latent space as explained above. If the reconstruction error is above the threshold, the mutation is applied directly in the parameter space using $\mathcal{N}(\theta_i, \Sigma_{\Theta})$.
        The reconstruction error threshold $\epsilon_{recn}$, is determined heuristically based on the average reconstruction error of all the points in the collection, achieved after the manifold learning phase.
        The pseudocode for region-based policy search is given in Algorithm~\ref{algo:mix_region}.
        %
        %
        This approach allows us to use mutations in the latent space in regions that are well known by the AE (i.e. low reconstruction error), and unconstrained mutations in the parameter space otherwise.
        Periodically adding solutions obtained via parameter space search helps reduce overfitting, thus making the AE generalise better.
        This can be regarded as a type of active learning based on model uncertainty~\cite{rakicevic2019active}.
        In the experiment section, when using the mixing strategy, we show its average mixing ratio (proportion of the total number of samples obtained in the parameter space).


\section{Evaluation}
\label{sec:evaluation}

    Our experimental evaluation~\footnote{The code for the proposed approach and evaluation environments is available online: \url{https://sites.google.com/view/policy-manifold-search}} aims to find the following insights:
    \\
    \textbf{Q1.} Are there any benefits of using the learned manifold for policy search, as opposed to the original parameter space? \\
    \textbf{Q2.} What is the effect of using the Jacobian of the decoder, for scaling the latent space sampling covariance matrix? \\
    \textbf{Q3.} Is the non-linear manifold learning necessary, or does a linear projection suffice? \\
    \textbf{Q4.} How does \method compare to state-of-the-art QD approaches?
    
    In order to answer Q1, we compare the performance of the proposed \method approach, to MAP-Elites (MAPE-Iso) which performs search in the original parameter space. 
    Within this analysis, we also consider two state-of-the-art baselines which are based on a notion of a manifold: 
    MAP-Elites via Elite Hypervolumes (MAPE-IsoLineDD)~\cite{vassiliades2018discovering} and Data-driven encoding MAP-Elites (DDE)~\cite{gaier2020discovering}.
    In this paper, we mainly focus on the parameter versus manifold exploration.
    While other work exists that scale MAP-Elites to Neuroevolution using more advanced search strategies in the parameter space~\cite{colas2020scaling}, we leave for future work the investigation on how these two families of approaches can be combined.
    
    Further, we perform an ablation study to address Q2 and Q3.
    The first aim is to demonstrate the importance of scaling the latent space sampling covariance matrix, using the decoder Jacobian.
    The alternative to this would be a latent space search, where the latent sampling covariance matrix $\Sigma_{Z}$ is determined based on the current ranges of the latent representations (PoMS-no-jacobian).
    The second aim is to examine the importance of non-linear manifold learning, using AE, as opposed to using a linear projection of the parameter space via PCA (PoMS-PCA) to obtain the latent space.
    
    Regarding Q4, we focus on how \method compares to MAPE-IsoLineDD and DDE.
    Additionally, we examined Diversity is All You Need (DIAYN) \cite{eysenbach2018diversity}, 
    as a task-conditioned policy approach.
    %
    However, preliminary tests showed that it is unable to scale to several thousand skills, and in the time required by MAP-Elites and \method to evaluate million episodes, DIAYN managed to evaluate just over 2000, without observing any promising results. Therefore, we decided to exclude this algorithm from our analysis.
    %
    %
    
	\textbf{Random baselines} are included as a `sanity check' to compare the proposed algorithm with random search. We evaluate two approaches: 
	(ps-uniform) where each policy network samples are drawn from a uniform distribution $\mathcal{U}(-1,1)$, and 
	(ps-glorot) in which each sample is initialised according to Xavier-Glorot scaled normal distribution, a popular network weight initialiser \cite{glorot2010understanding}.
	
    %
    \begin{figure}[t!]
            \centering
            \subfloat[Bipedal-\\Walker\label{fig:envs:biped-walk}]
            {\includegraphics[width=0.115\textwidth]{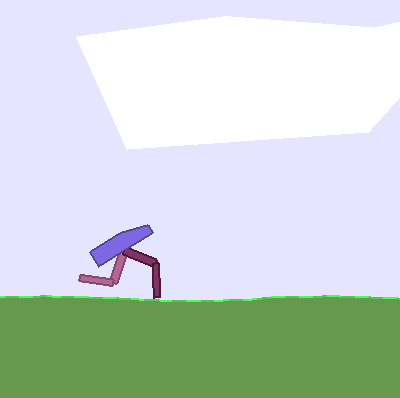}} 
            \hfill
            \subfloat[Bipedal-\\Kicker\label{fig:envs:biped-kick}]
            {\includegraphics[width=0.115\textwidth]{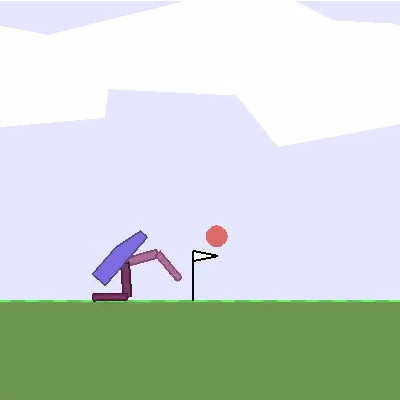}}
            \hfill
            \subfloat[2D-Striker\label{fig:envs:striker}]
            {\includegraphics[width=0.115\textwidth]{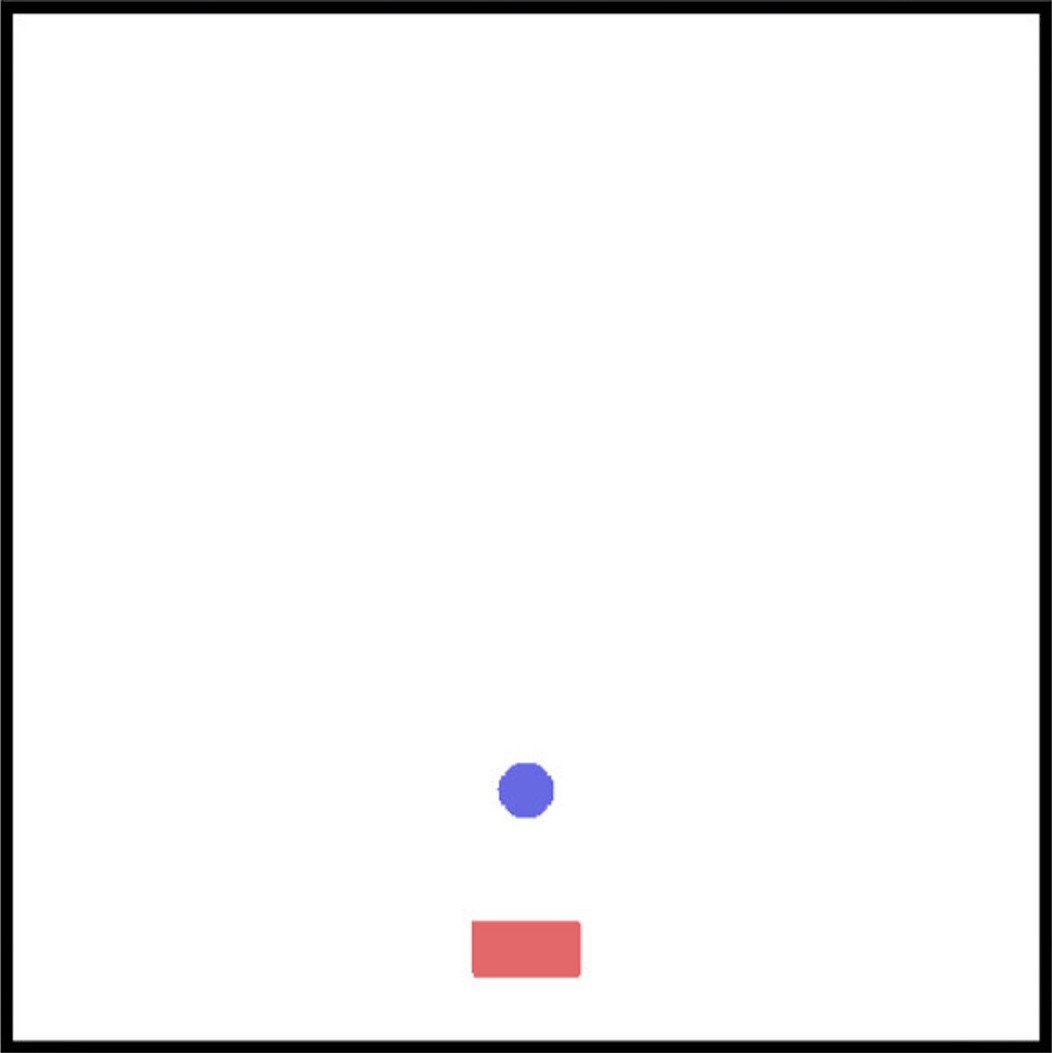}} 
            \hfill
            \subfloat[Panda-\\Striker\label{fig:envs:panda}]
            {\includegraphics[width=0.115\textwidth]{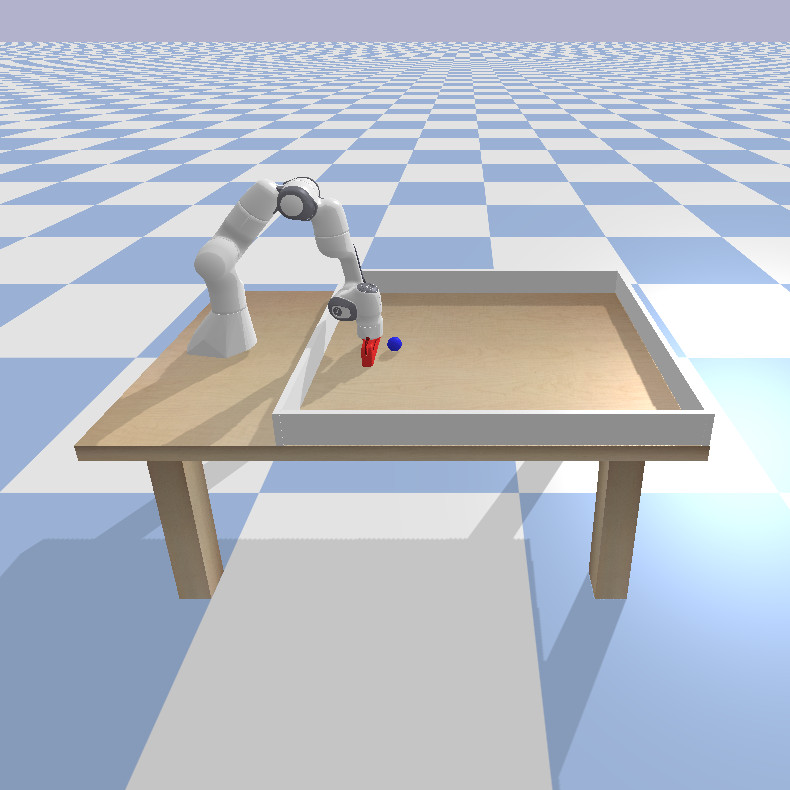}} 
            \caption{Screenshots of the continuous control environments used for experimental evaluation. The environments (a-c) are implemented in Box2D \cite{catto2011box2d} and (d) in PyBullet \cite{coumans2016pybullet}.}
            \label{fig:envs}
    \end{figure}

    \begin{figure*}[tb!]
        \centering
        {\includegraphics[width=.98\textwidth]{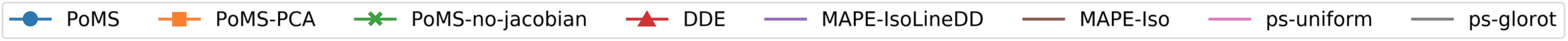}} 
        \\
        \subfloat[Bipedal-Walker]{\includegraphics[width=0.245\textwidth]
        {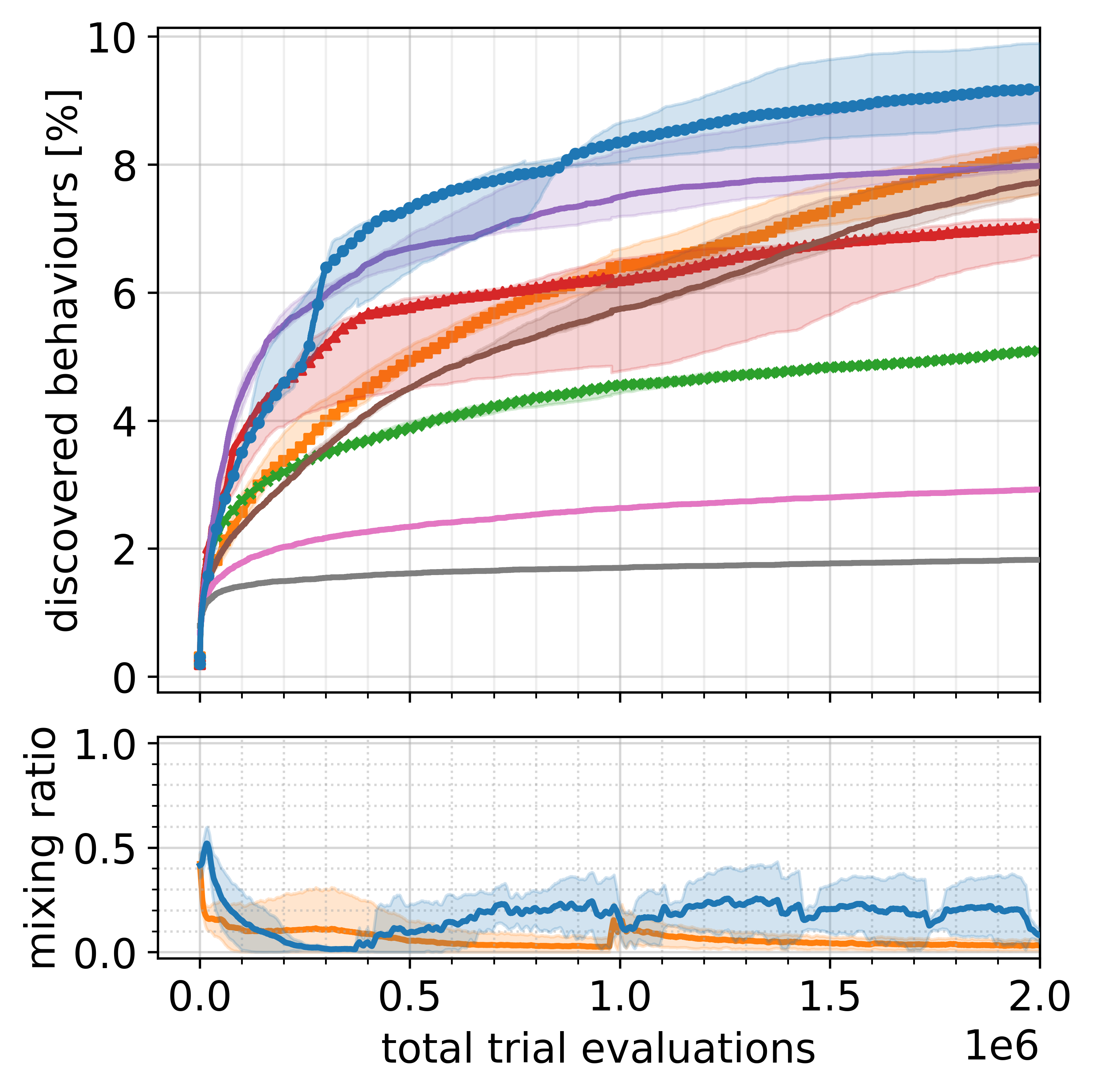}
        \label{fig:res_plots:biped_walk}} 
        \subfloat[Bipedal-Kicker]{\includegraphics[width=0.245\textwidth]
        {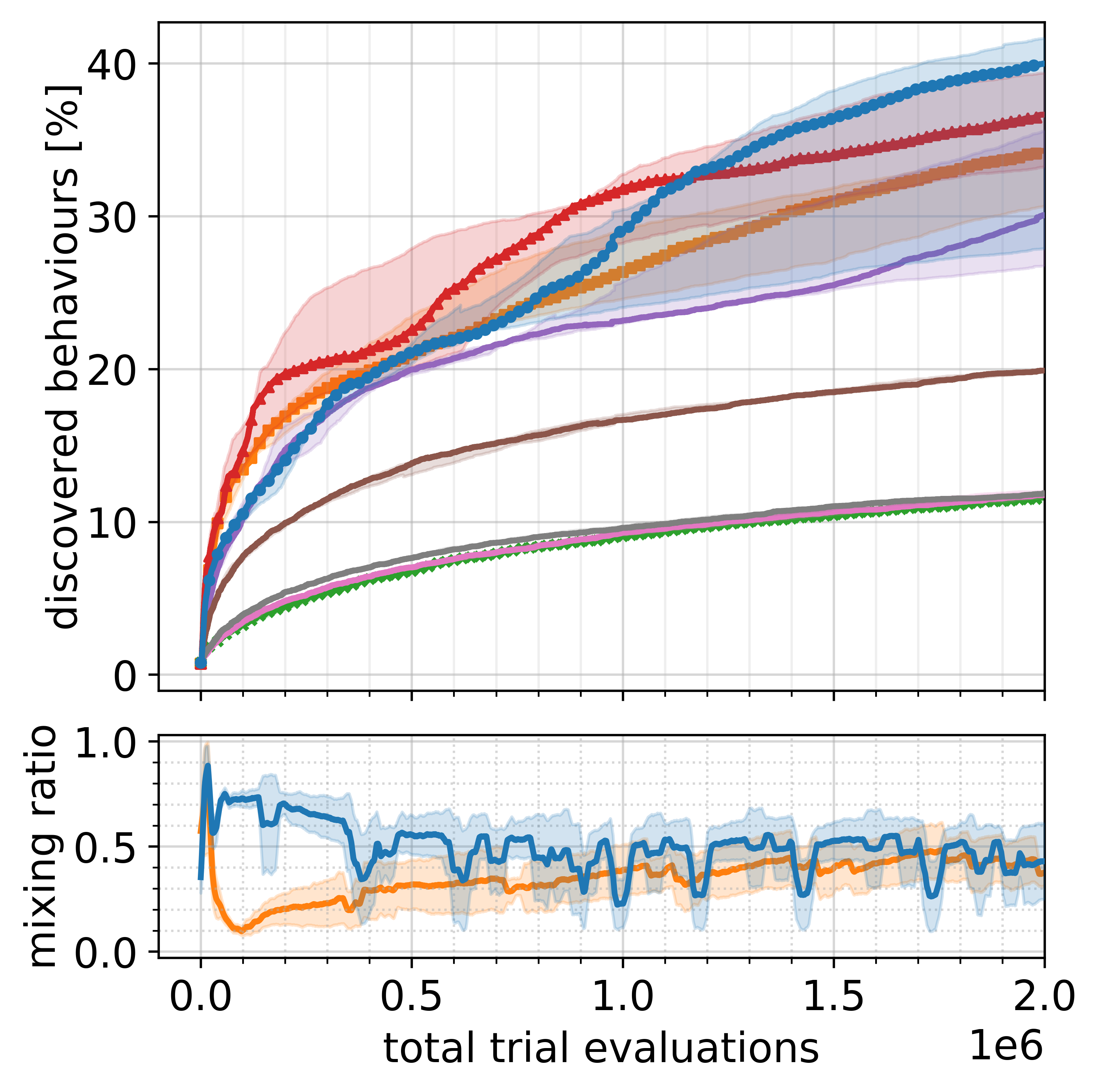}
        \label{fig:res_plots:biped_kick}} 
        \subfloat[2D-Striker]{\includegraphics[width=0.245\textwidth]
        {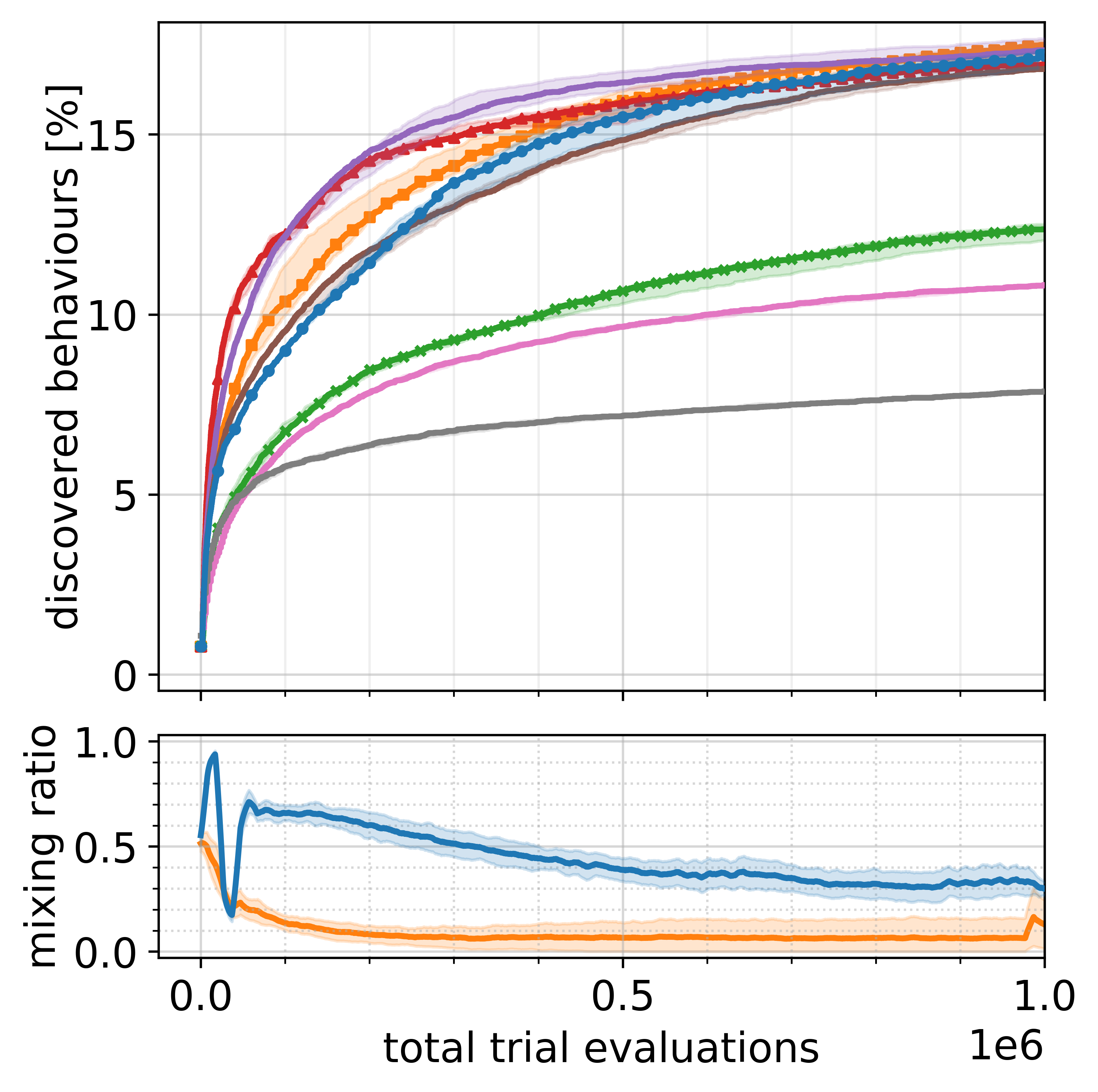}
        \label{fig:res_plots:striker}} 
        \subfloat[Panda-Striker]{\includegraphics[width=0.245\textwidth]
        {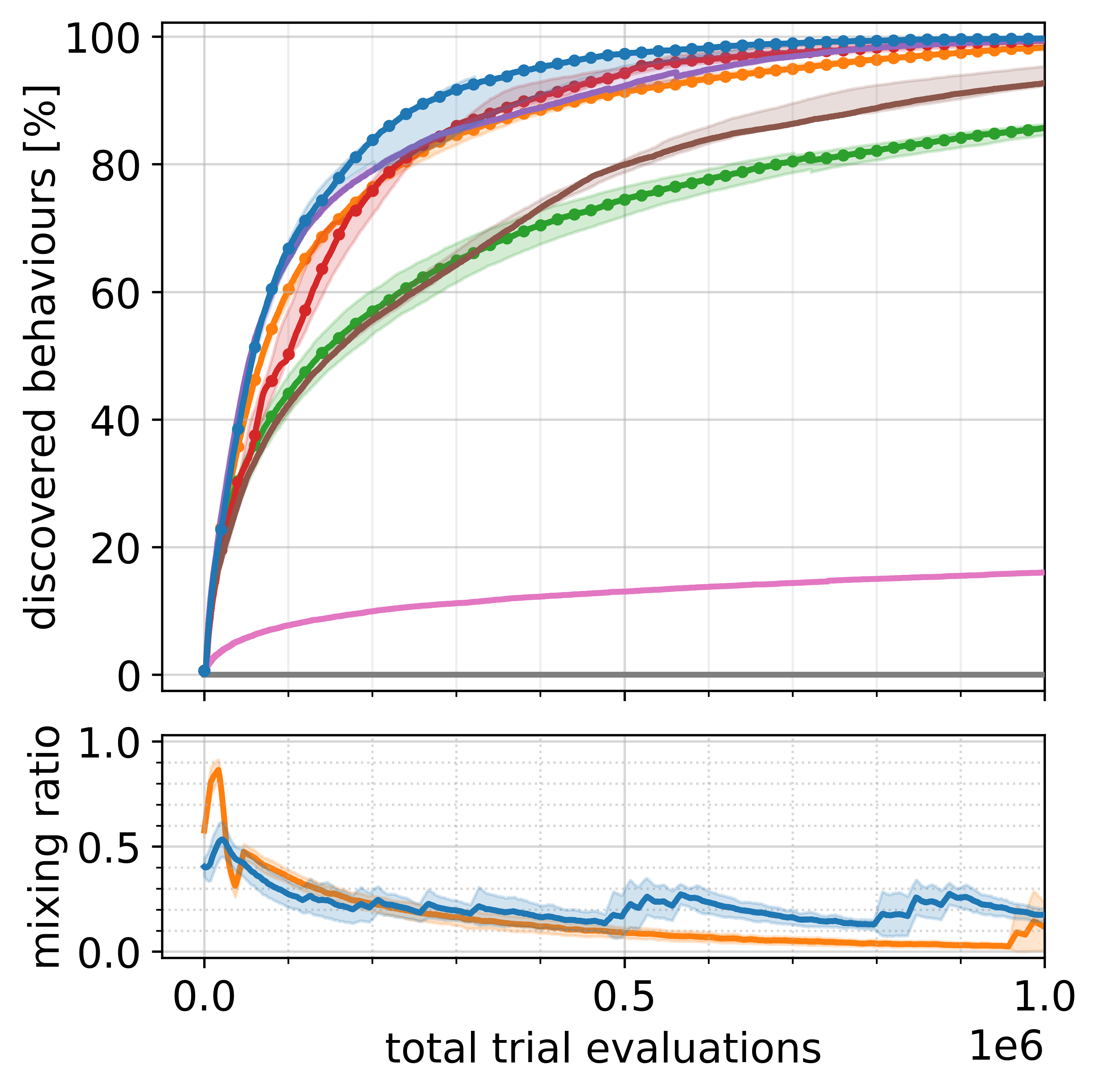}
        \label{fig:res_plots:panda_striker}}
    \caption{Behaviour coverage and mixing-ratio plots achieved by the compared methods, for four continuous control environments. The markers on the lines of approaches using a latent representation, correspond to latent representation updates.
    }
    \label{fig:res_plots}
    \end{figure*}
    \subsection{Experiments}
        We evaluate the methods on four deterministic simulated environments (shown in Fig. \ref{fig:envs}). 
        The agent acting in the environments is controlled by a neural network parameterised policy, whose parameters are generated by the policy search method and stored in the policy collection.
        The policy network takes the full observation vector as input, and outputs a desired action vector (joint torques or velocities).
        The architecture of the policy network is kept the same across all experiments, and is based on the Proximal Policy Optimisation \cite{schulman2017proximal} policy implementation for continuous control tasks. 
        It is as a fully-connected neural network with two hidden layers of 32 neurons each, with $tanh$ activation functions. The output layer activation function is linear. The input and output size vary depending on the observation and action vector sizes, which are specific to each environment.
        The environments used are:
        %
        %
        %
        %
        %
    	
        \textbf{Bipedal-Walker [observation 26D, action 4D, collection size 50000]}
        is a standard OpenAI gym \cite{brockman2016openai} environment (Fig. \ref{fig:envs:biped-walk}).
        The original observation vector has 24 elements, which include the robot hull angle, horizontal, vertical and angular velocities, joints angles and angular velocities, legs-ground contact information, and 10 lidar rangefinder measurements. 
        We also added the absolute coordinates of the robot hull, thus creating a 26D observation vector.
        The 4D action vector is unaltered and provides torques for each of the leg joints.
        At the start of the episode, the robot is placed in the middle of the terrain so it can walk either forward or backwards, in as many diverse ways as possible.
        The episode is limited to 500 steps.
        The policy collection is a 4D grid populated using the behaviour descriptor based on the agent's absolute positions and leg-ground contacts during the episode, where each dimension index is: 
        D1: average hull y-coordinate, limited to [4.5, 6.2] range and discretised into 5 bins;
        D2: final hull x-coordinate, 100 bins spanning the terrain length;
        D3-4: proportion of time left and right legs spent in contact with ground, respectively, normalized to [0,~1] range discretised into 10 bins each.

\begin{figure*}[t!]
        \centering
        \subfloat[PoMS]{\includegraphics[width=0.3\linewidth]
        {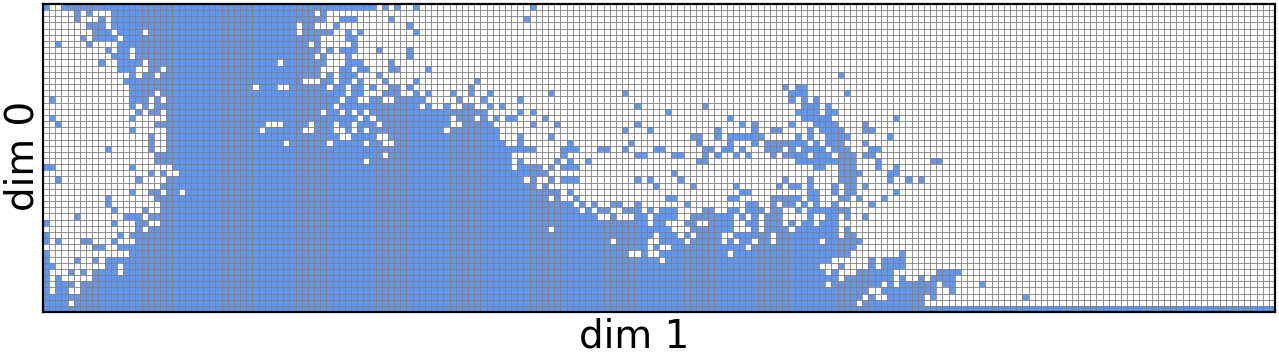}
        \label{fig:collection:poms}} 
        \hfill
        \subfloat[PoMS-PCA]{\includegraphics[width=0.3\linewidth]
        {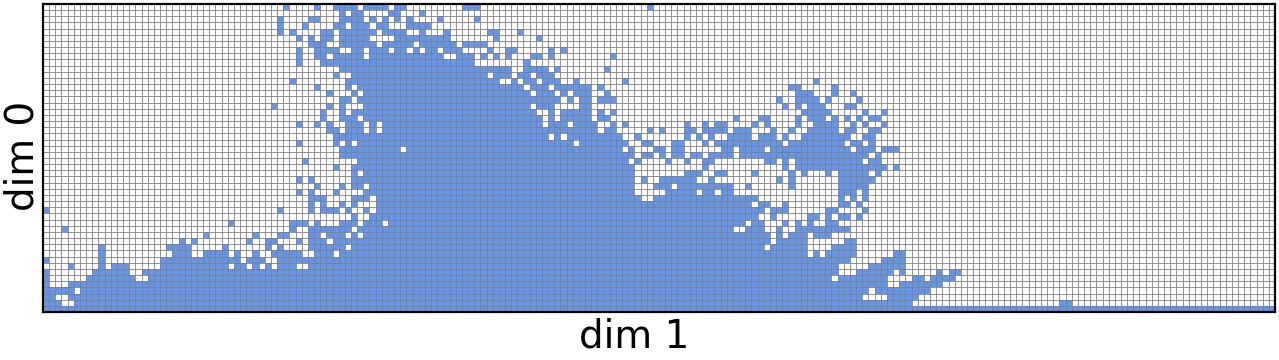}
        \label{fig:collection:poms_pca}} 
        \hfill
        \subfloat[PoMS-no-jacobian]{\includegraphics[width=0.3\linewidth]
        {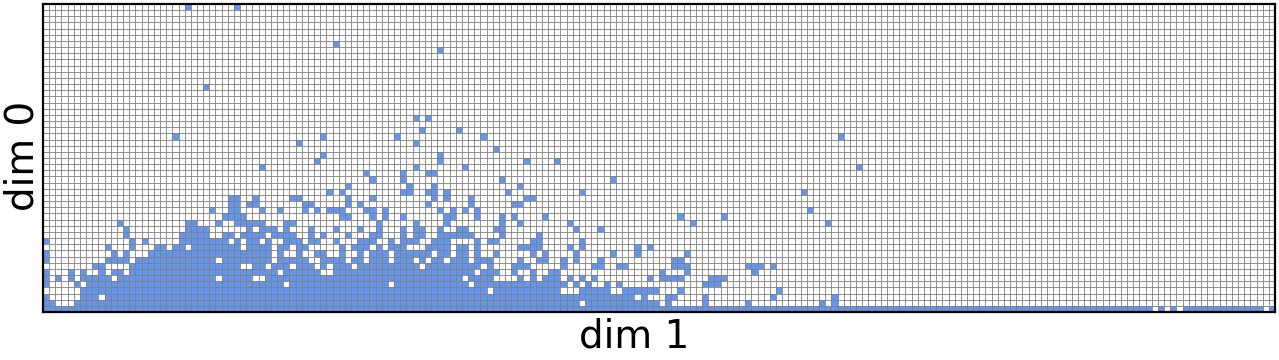}
        \label{fig:collection:poms_no_jacobian}} 
        \\
        \subfloat[MAPE-IsoLineDD]{\includegraphics[width=0.3\linewidth]
        {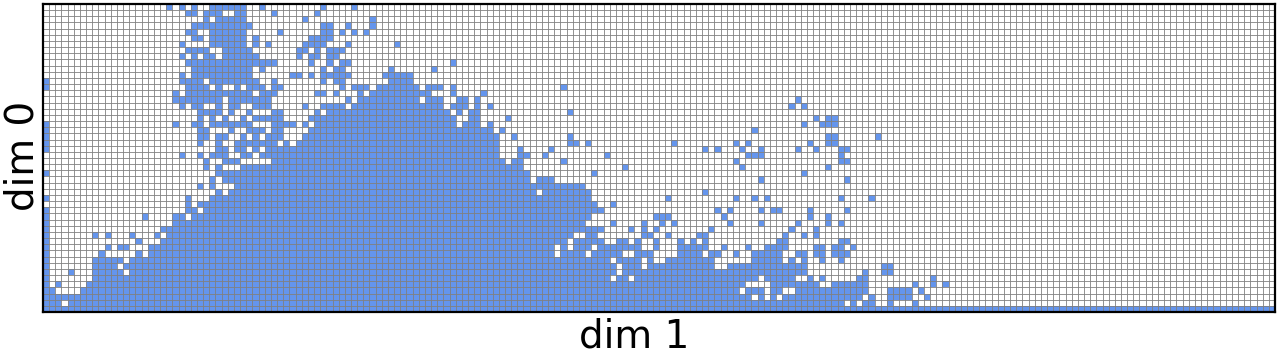}
        \label{fig:collection:mape-isolinedd}} 
        \hfill
        \subfloat[DDE]{\includegraphics[width=0.3\linewidth]
        {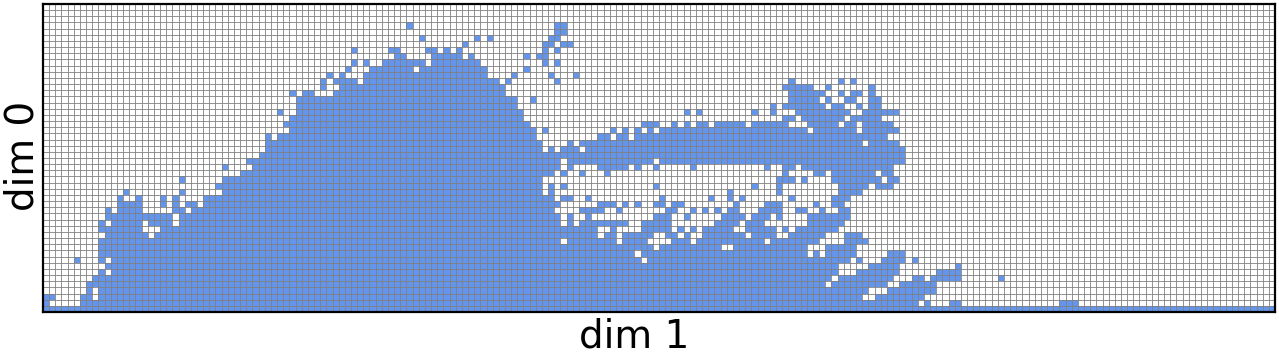}
        \label{fig:collection:dde}} 
        \hfill
        \subfloat[MAPE-Iso]{\includegraphics[width=0.3\linewidth]
        {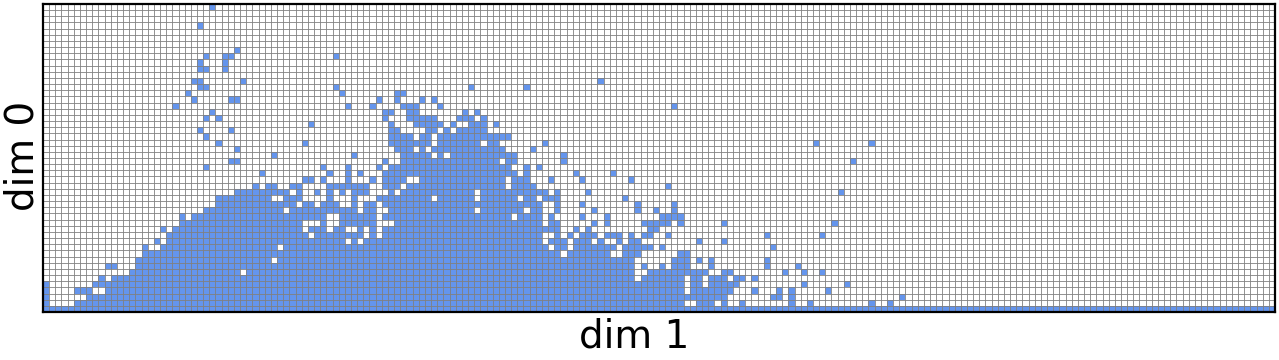}
        \label{fig:collection:mape-iso}}
    \caption{
    Policy collections showing the behaviour coverage achieved by the compared methods for the Bipedal-Kicker.
    }
    \label{fig:collection}
    \end{figure*}

        \textbf{Bipedal-Kicker [observation 30D, action 4D, collection size 10000]}
        extends the Bipedal-Walker task by adding a ball (Fig. \ref{fig:envs:biped-kick}). Therefore, the observation vector is extended with the ball x, y position and velocities, making it 30D. The action output is the same.
        Since the goal is to have a diversity of ball ballistic trajectories, we make the terrain flat to avoid biasing the outcomes to local valleys.
        In order to facilitate kicking, as the agent does not have a foot, at the start of the episode the ball is dropped from a small height so the agent can hit it.
        The agent is allowed to move for 100 timesteps and then stops to avoid multiple kicks, while the ball moves until it stops due to the damping effects.
        The behaviour descriptor, used for the 2D grid policy collection (Fig. \ref{fig:collection}), is based on the ball trajectory, as a usual way of defining a 2D ballistic trajectory:
        D1: final ball x-coordinate, in the right half of the terrain divided into 200 bins;
        D2: maximum ball y-coordinate achieved during the episode, limited to [4, 7] range and discretised into 50 bins.

        \textbf{2D-Striker [observation 14D, action 3D, collection size\\15300]} 
        is a bounded air-hockey-like environment implemented in Box2D \cite{catto2011box2d},
        with the goal of controlling the striker to hit the puck so it lands on as many diverse positions as possible (Fig. \ref{fig:envs:striker}).
        The arena is bounded by four walls to the size of 100x100 units, created to be proportional to the striker size (5x2.5 units). The puck has a radius of 2.5 units.
        The 14D observation vector, consists of the striker's x, y position and angle $\phi$, the puck's x, y position, as well as their corresponding velocities, and puck-wall distances for each of the walls.
        The 3D action vector controls the striker's x, y and angular velocities.
        Similarly to Bipedal-Kicker, the agent is allowed to act 100 steps before the actions are set to 0, in order to have only one interaction with the puck per episode, while the puck moves until it stops due to damping effects.
        The policy collection is a 3D grid, with the indices corresponding to: 
        D1-2: final x, y position of the puck, with 30 bins per dimension;
        D3: index wall(s) the puck bounced off during the episode, and has 17 possible values (no wall, south, east, north, west, and 12 additional second order combinations).

        \textbf{Panda-Striker [observation 24D, action 7D, collection size 15300]} 
        has the same goal, policy collection and behaviour descriptor as 2D-Striker.
        However, here the striker is attached to the end-effector of a 7-DOF Franka-Panda arm implemented in PyBullet~\cite{coumans2016pybullet}, which is used to hit the ball on a bounded table (Fig. \ref{fig:envs:panda}).
        The 24D observation vector contains the 7 joint positions and velocities, cartesian position and orientation of the end-effector and the ball's position and velocities.
        The 7D action vector contains joint velocity controls for moving the robot arm.

    	In contrast to Bipedal-Walker, the Bipedal-Kicker and both Striker environments contain a ball, which is an external object manipulated by the agent. This adds complexity to the task as certain elements of the observation vector can vary independently of the agent's actions. 
    	Another distinction is that Striker environments are bounded, while Bipedal ones are not.
    	This leads to certain elements of the observation vector having different scales. 
    	Usually they can be normalised, but in unbounded environments this is not straightforward.
    	We investigate the influence of these environment differences on the behaviour diversity via the evaluation metric.

    \subsection{Metric}
	    %
        %
        We use \textit{behaviour coverage} as a metric to quantify the behaviour diversity, in order to assess a search method's performance \cite{cully2017quality}.
        It quantitatively describes the number of distinct behaviours discovered by the algorithm, as a percentage of the policy collection filled.
        For the experiments to have statistical significance, 5 runs are executed with different parameter initialisation seeds.
        We show the median of the 5 seeds, 25th and 75th percentile.
        %
        The x-axis of the behaviour coverage plots shows the total cumulative number of episode rollouts to have a fair comparison.
    %
    %
    %
    %


\section{Results}
\label{sec:results}

    Below, we discuss the behaviour coverage achieved by the compared methods (Fig. ~\ref{fig:res_plots}), in the context of questions from Sec. \ref{sec:evaluation}.
    Implementation details of these methods are given in Appendix~\ref{appendix:implementation}.
    
    \textbf{A1. Learned latent space vs original parameter space.}\\
    The first conclusion we note from the experiments, is that the standard MAP-Elites algorithm (MAPE-Iso) is competitive in high-dimensional parameter space problems, which has not been sufficiently investigated in previous work.
    Comparing the proposed \method approach with MAPE-Iso, \method systematically achieves higher behaviour coverage across the tasks.
    This is especially evident with Bipedal-Kicker, where the relationship between the observation vector and the action outputs, which depend on the policy parameterisation, is highly complex.
    This could be attributed to the unbounded nature of the environment and the presence of an additional object.
    The 2D-Striker case is an exception as all methods converge to the same performance (Fig. \ref{fig:envs:striker}).
    This result can be explained by the simplicity of the task, as the policy outputs can directly influence the planar movement of the striker, and by extension the puck, while such connection is more complex within Bipedal environments and to some extent with the Panda arm.
    Even though MAPE-IsoLineDD operates in the parameter space, it uses the hypervolume of elites for search, which can be related to the notion of a manifold, thus improving its efficiency over MAPE-Iso. 
    However, besides the Striker tasks where it reaches equal asymptotic performance, it converges to a lower behaviour coverage compared to \method.
    By definition of the line mutation, the MAPE-IsoLineDD usually performs well when the hypervolume of elites is convex. When this assumption does not hold, a more involved transformation is needed which \method realises via manifold learning.
    DDE uses the learned latent representation indirectly by exploiting the reconstruction inaccuracies through its mutation operator, which on average achieves higher behaviour coverage than direct policy space search, such as MAPE-Iso. 
    Still, accounting for the decoder transformation and having a more structured latent sampling as in \method enables better performance.
    To quantitatively validate our findings, we estimate the statistical significance of the actual improvements that the methods using the latent space for policy search have over the parameter space based methods.
    We use the Mann Whitney Wilcoxon test, a non-parametric test whose alternative hypothesis is that values in one sample are more likely to be larger than the values in the other sample, and this takes into account the consistency over multiple seeds at convergence.
    This statistic is calculated for each of the latent space based methods (\method, MAPE-IsoLineDD and DDE) in comparison with MAPE-Iso.
    We report the highest $p$-value of the three pairs, in each of the environments:
    Bipedal-Walker ($p$=0.25),     
    Bipedal-Kicker ($p$=0.009),    
    2D-Striker ($p$=0.21)          
    and Panda-Striker ($p$=0.009). 
    There are two reasons explaining the cases which are inconclusive.
    In the 2D-Striker case, this is due to all approaches converging to the same value.
    In the Bipedal-Walker case, the approaches that do not explicitly learn a manifold struggle with the non linear aspects of the environment, while \method maintains statistically significant results ($p$=0.047).
    These findings validate the benefits of using a learned latent representation of the policy parameters for policy search.
    
    \textbf{A2. Contribution of decoder Jacobian scaling.}
    By comparing \method and PoMS-no-jacobian, we can see that accounting for the decoder transformation of the latent space samples, via Jacobian scaling of their covariance matrix, is crucial for the algorithm's performance.
    As hypothesised, simple latent perturbation and reconstruction done in PoMS-no-jacobian, leads to a search that performs similarly to random search (Fig. \ref{fig:res_plots:biped_kick}, \ref{fig:res_plots:striker}).
    
    \textbf{A3. Linear vs non-linear representations.}
    The difference in performance between \method and PoMS-PCA, speaks mostly about the intrinsic complexity of the given task control problem. 
    Environments in which locomotion is involved have an intrinsic non-linearity in the mapping of the policy outputs and actual motions contained in the observation vector, which is used to determine behaviour descriptors.
    This can explain the performance gap between \method and PoMS-PCA in both Bipedal environments (Fig.~\ref{fig:res_plots:biped_walk}, \ref{fig:res_plots:biped_kick}). 
    
    \textbf{A4. State-of-the-art performance comparison.}
    The methods with consistently highest behaviour coverage are PoMS, PoMS-PCA, DDE and MAPE-IsoLineDD, which are all using a notion of a manifold, by focusing the search in the hypervolume of elites or the learned manifold.
    This further solidifies claims from A1.
    Even though DDE and MAPE-IsoLineDD were not originally used with neural network parameterised controllers, they perform well in high-dimensional parameter spaces.
    As we increase the environment complexity, by adding a Panda arm to Striker, and considering the Bipedal agent, we can observe that the proposed PoMS, maintains a high behaviour coverage, while other state-of-the-art methods are not as robust.
    This drop in performance is higher in purely parameter space search methods such as MAPE-Iso, ps-uniform and ps-normal.
    \method outperforms the next best state-of-the-art approach by 
    a relative increase of the median coverage of $15\%$ and $9\%$ respectively for Bipedal-Walker and Bipedal-Kicker environments, while it is on-par in both Striker environments at convergence.
    A qualitative comparison of the achieved behaviour coverage at convergence, for each of the compared methods is shown in Fig. \ref{fig:collection}.
    %
    
    \textbf{Mixing ratio}
    Below the corresponding behaviour coverage graphs in Fig. \ref{fig:res_plots}, we show the mixing ratios for \method and PoMS-PCA.
    The mixing ratio represents the averaged ratio of samples generated in the latent, versus the parameter space, during one search iteration.
    Mixing ratio of 1 means that all the policy parameters are sampled directly in the parameter space, while 0 means that all are sampled in the latent space (using the scaled covariance matrix) and then reconstructed.
    The first loop of the algorithm starts with a mixing ratio of 0.5 and subsequently the ratio changes based on the mean reconstruction error, as explained in Section~\ref{sec:method}.
    From the mixing ratios we can see that in the beginning, there is usually a spike towards parameter space sampling.
    This is due to the fact that initially there are many points with a high reconstruction error, because the AE is trained on a small amount of data, and needs more diversity - thus it `explores'.
    The mixing ratio slowly decreases in favor of the latent space samples, with several salient `dips' which loosely correlate to jumps in behaviour discovery. This can be interpreted as the algorithm `exploiting' the learned latent representation. 
    This is not evident with PCA as its representations tend to be more rigid and do not change often with new data. 
    

\vspace{-0.13cm}
    
\section{Conclusion}
\label{sec:conclusion}
    \vspace{-0.05cm}
    In this paper, we proposed the Policy Manifold Search algorithm, inspired by the manifold hypothesis, that aims to discover a collection of policies with diverse behaviours by performing search in the learned manifold embedded in the policy parameter space.
    %
    %
    We learn this manifold from a collection of policy parameters maintained according to the MAP-Elites framework, and use the manifold to generate novel solutions, thus augmenting the collection. 
    Experimental evaluations of \method validate the benefits of using a learned manifold, coupled with the Jacobian of the decoder for guiding the search, to discover larger collections of diverse policies compared to the baselines. 
    The benefits of considering the manifold hypothesis were further reinforced by examining two state-of-the-art methods which rely on this notion as well, and achieve good performance.
    %
    %
    
    In future work, we plan to further investigate the structure of the learned manifold and how it can be formed without using domain knowledge through behaviour descriptors.
    Another interesting line of work we plan to pursue is extending this approach to full network graph encoding, rather than just weight parameters.


\begin{acks}
    This work was supported by the Imperial College London President's PhD Scholarship, and Engineering and Physical Sciences Research Council (EPSRC) grant EP/V006673/1 project REcoVER.
    %
    %
\end{acks}


\appendix 

\section{Autoencoder Training Details}
\label{appendix:training}
    During the \textit{parameter manifold learning} phase of PoMS, the AE is trained by minimising the reconstruction loss $\mathcal{L}_{AE}$ in order to find the optimal AE parameters ($\xi$).
    We use the Adam optimiser with: $\beta_1=0.9$, $\beta_2=0.999$, $\varepsilon=10^{-8}$ and learning rate of $10^{-5}$, for $2 \cdot 10^{4}$ epochs, with batch size of 64. 
    The training hyperparameters are the same in all experiments.
    To improve the robustness of the optimiser, we reset the momentum variables at every loop.
    Moreover, 30\% of each batch is used as a test set for early stopping of the training. If the slope of the line fitted to the last 100 test set values is larger than $10^{-5}$, the training is stopped. We found that this improves the generalisation of the AE and reduces training time.

\section{Implementation Details}
\label{appendix:implementation}
    All algorithms based on MAP-Elites (PoMS versions, MAPE-Iso, MAPE-IsoLineDD and DDE) run 100 iterations of MAP-Elites with a budget of 200 samples. 
    The policy collection is initialised by drawing 2000 policy samples from a uniform distribution, as usually done in the MAP-Elites literature \cite{cully2017quality}.
    The ps-uniform and ps-glorot methods run for the same total amount of samples as other algorithms, while the progress is displayed every 2000 samples.
    \\         
    \textbf{PoMS}
        has three main hyperparameters that need to be tuned: AE architecture, latent space dimension (LD) and $\Sigma_{\Theta}$. 
        The AE is symmetric, i.e. both encoder and decoder have one hidden layer with ELU activations, and we vary the number of hidden nodes (HD). The activation function of the bottleneck layer forming the latent space is linear, same as for the output layer of the decoder.
        The hyperparameter values used in the experiments are:
        \\
        \begin{tabular}{|p{4cm}|p{1cm}|p{1cm}|p{1cm}|}
            \hline
            \textbf{Environment}        & \textbf{HD}   & \textbf{LD}   & $\bm{\Sigma}_{\Theta}$ \\
            \hline
            \hline
            Bipedal-Walker              & 100           & 50            & 0.1    \\
            \hline
            Bipedal-Kicker              & 100           & 100           & 0.01   \\
            \hline
            2D-Striker                  & 100           & 50            & 0.1    \\
            \hline
            Panda-Striker               & 100           & 50            & 0.5    \\
        \hline
        \end{tabular}
    \\\\
    \textbf{PoMS-no-jacobian}
        keeps the same AE architectures as PoMS, in order to have an appropriate ablation study.
        The $\Sigma_{Z}$ is not fixed, rather, dynamically updated based on the current latent space parameter ranges, per latent dimension $\bm{r}_{Z}$, in order to scale a unit covariance matrix $\Sigma_{Z} = \bm{r}_{Z}^T\mathbb{I}$.
        In this way, we give importance of each of the latent dimensions based on their spread. 
        As we can see from the results, two issues arise with this approach:
        (i) range does not equal importance (solution density),
        (ii) applying the inverse transformation applies an additional distortion which can lead to undesirable values, because
        $\theta' \neq f_{D}(z')$, where $ z' \sim \mathcal{N}(\mu_Z, \Sigma_{Z})$ and $ \theta' \sim \mathcal{N}(f_{D}(\mu_Z), \Sigma_{\Theta})$, even if $\Sigma_{\Theta} = \Sigma_{Z} = \mathbb{I}$.
    \\
    \textbf{PoMS-PCA}
        requires LD and $\Sigma_{\Theta}$ as hyperparameters.
        These values are kept the same as with PoMS, for a proper ablation study.
    \\
    \textbf{DDE}
        hyperparameters consist of the AE architecture and mutation operator specific hyperparameters.
        The former is kept the same as with PoMS, for each of the experiments, while the latter are as in the original paper \cite{gaier2020discovering}.
        Instead of running a fixed window for the multi-armed bandit upper confidence bound operator selector, we maintain a moving average.
    \\
    \textbf{MAPE-IsoLineDD}
        has two hyperparameters related to the weighing of the isometric and directional components of the mutation operator, and they are kept the same as in the original paper~\cite{vassiliades2018discovering}.
    \\
    \textbf{MAPE-Iso}
        needs only $\Sigma_{\Theta}$, which we set to $\Sigma_{\Theta}$=0.1 as this achieved the best performance for MAPE-Iso across the experiments.

\section{Derivation of Jacobian scaling}
\label{appendix:jacobian}
    Let us assume $\theta \in \Theta \subset \mathbb{R}^P$ and $z \in Z \subset \mathbb{R}^M$ to be Gaussians in the parameter and latent space respectively, such that $\theta \sim \mathcal{N}(\mu_{\Theta}, \Sigma_{\Theta})$ and $z \sim \mathcal{N}(\mu_Z, \Sigma_Z)$.
    We further assume a non-linear, vector-valued function $f_{D}: \mathbb{R}^M \rightarrow \mathbb{R}^P$ which maps the latent space to the original parameter space.
    Then, we can get a linear approximation  $\hat{f}_{D}$ in a point $\mu_Z$ by performing a first-order Taylor expansion:
    \begin{equation}
    \label{eq:taylor}
    \begin{split} 
        \hat{f}_{D}(z)  & \approx f_{D}(\mu_z) + 
                                 \sum_i^M \frac{\partial f_{D}(z)}{\partial z_i} \biggr\rvert_{z=\mu_{Z}} 
                                 (z_i - \mu_{Zi}) \\
                        & = f_{D}(\mu_Z) + 
                                \begin{bmatrix}
                                    \nabla f_{D}(z)_1\\
                                    \nabla f_{D}(z)_2\\
                                    ...\\
                                    \nabla f_{D}(z)_P
                                \end{bmatrix}
                                (z - \mu_{Z}) \\
                        & = \mu_{\Theta} + \mathbf{J}_{D}(z - \mu_{Z})
    \end{split}
    \end{equation}
    where $\mathbf{J}_{D}$ is the Jacobian matrix of $f_{D}$ at $\mu_Z$:
    \begin{equation}
    \label{eq:jacobian}
        \mathbf{J}_{D}(\mu_Z) = J_{D}(\mu_Z)_{ij} = \frac{\partial f_{D}(\mu_Z)_i}{\partial \mu_{Z_j}}
    \end{equation}
    where indices $i$ and $j$ refer to the corresponding elements of the reconstructed or latent parameter vector, respectively.
    Further, if $\hat{\theta} = \hat{f}_{D}(z)$ its expected value can be obtained as:
    \begin{equation}
    \begin{aligned}
        \mathbb{E}[\hat{\theta}] & = \mathbb{E} \left [ \mu_{\Theta} + \mathbf{J}_{D}(z - \mu_{Z}) \right ]
                        = \mathbb{E}[\mu_{\Theta}] + \mathbb{E} \left [\mathbf{J}_{D}(z - \mu_{Z}) \right ] \\
                        & \text{(expected values of a sum is the sum of expected values)} \\
                        & = \mu_{\Theta} + \mathbb{E} \left [\mathbf{J}_{D} z \right ]  - \mathbb{E} \left [\mathbf{J}_{D} \mu_{Z} \right ] \\ 
                        & \text{(expectation of a constant is a constant)} \\    
                        & = \mu_{\Theta} + \mathbf{J}_{D} \mathbb{E} [z] - \mathbf{J}_{D} \mathbb{E} [\mu_{Z}]
                        = \mu_{\Theta} + \mathbf{J}_{D} \mu_{Z} - \mathbf{J}_{D} \mu_{Z} \\ 
                        & = \mu_{\Theta}   
    \end{aligned}
    \end{equation}
    We can obtain $\Sigma_{\Theta}$ based on $\Sigma_Z$. We start with the standard equation for covariance:
    \begin{equation}
    \begin{aligned}
        \Sigma_{\Theta} & = \mathbb{E} \left [(\hat{\theta} - \mathbb{E}[\hat{\theta}])(\hat{\theta} -                               \mathbb{E}[\hat{\theta}])^T \right ] \\
                        & = \mathbb{E} \left [(\mu_{\Theta} + \mathbf{J}_{D}(z - \mu_{Z}) - \mu_{\Theta})
                            (\mu_{\Theta} + \mathbf{J}_{D}(z - \mu_{Z}) - \mu_{\Theta})^T \right ] \\
                        & \text{(using Equations \ref{eq:taylor} and \ref{eq:jacobian})} \\
                        & = \mathbb{E} \left [(\mathbf{J}_{D}(z - \mu_{Z}))
                            (\mathbf{J}_{D}(z - \mu_{Z}))^T \right ] \\
                        & = \mathbb{E} \left [ \mathbf{J}_{D}(z - \mu_{Z})
                            (z - \mu_{Z})^T \mathbf{J}_{D}^T \right ] 
                        \\
                        & 
                        = \mathbf{J}_{D} \mathbb{E} \left [ (z - \mu_{Z})
                            (z - \mu_{Z})^T \right ] \mathbf{J}_{D}^T \\
                        & \text{(covariance definition for z)} \\
                        & = \mathbf{J}_{D} \Sigma_Z \mathbf{J}_{D}^T
    \end{aligned}
    \end{equation}
    By rearranging the previous equation we finally get:
$
        \Sigma_Z = \mathbf{J}_{D}^T \Sigma_{\Theta} \mathbf{J}_{D}
$


\bibliographystyle{ACM-Reference-Format}
\bibliography{references} 


\end{document}